\documentclass[a4paper]{article}
\usepackage{ifthen}
\usepackage{url}
\usepackage{tabularx}
\usepackage{multirow}
\usepackage{tabularx}
\usepackage{graphicx}
\usepackage{amssymb}
\usepackage{CJKutf8} 
\usepackage{INTERSPEECH2021}
\newboolean{blind}
\setboolean{blind}{true} 
\title{Quantity Convergence, Quality Divergence: Disentangling Fluency and Accuracy in L2 Mandarin Prosody}
\name{Yuqi Shi$^1$, Hao Yang$^2$, Xiyao Lu$^1$, Jinsong Zhang$^1$}

\address{
  $^1$School of Information Science, Beijing Language and Culture University, Beijing, China \\
  $^2$School of Aeronautics and Astronautics, Sun Yat-sen University, Shenzhen, China
}
\email{shiyuqi5577@outlook.com, howyoung80@163.com, luxiyao1999@163.com, jinsong.zhang@blcu.edu.cn}
\begin{document}

\maketitle
\begin{abstract}
While second language (L2) learners may acquire target syntactic word order, mapping this syntax onto appropriate prosodic structures remains a persistent challenge. This study investigates the fossilization and stability of the L2 syntax-prosody interface by comparing 67 native Mandarin speakers with 67 Vietnamese learners  using the BLCU-SAIT corpus. By integrating C-ToBI boundary annotation with Dependency Grammar analysis, we examined both the quantity of prosodic boundaries and their mapping to syntactic relations.
Results reveal a non-linear acquisition: although high-proficiency learners (VNH) converge to the native baseline in boundary \textit{quantity} at the Major Phrase level(B3), their structural \textit{mapping} significantly diverges. Specifically, VNH demote the prosodic boundary at the Subject-Verb (SBV) interface (Major Phrase B3 → Prosodic Word B1), while erroneously promoting the boundary at the Verb-Object (VOB) interface (Prosodic Word B1 → Major Phrase B3). This strategy allows learners to maintain high long phrasal output at the expense of structural accuracy. This results in a distorted prosodic hierarchy where the native pattern is inverted.
\end{abstract}
\noindent\textbf{Index Terms}: Prosody boundary, Second Language Acquisition, Prosody-Syntax interface, Dependency grammar

\section{Introduction}
Prosodic structure acts as a critical interface between the continuous acoustic stream of speech and the discrete hierarchical structure of syntax, aiding listeners in parsing syntactic constituents and resolving structural ambiguities~\cite{Selkirk1984,Watson2004,Beach1991,Cooper1980}. In Mandarin Chinese, the alignment between prosodic boundaries and syntactic structure is governed by a complex interaction of constraints, often analyzed under the framework of the syntax-prosody interface~\cite{feng2019prosodic,Selkirk1986,Nespor2012,interface2011}. Generally, native speakers exhibit a sensitivity to syntactic cohesion: they tend to produce stronger prosodic boundaries at loose syntactic junctures while maintaining tighter prosodic connectivity within cohesive structures\cite{Gao2018,Gao2020,feng2019prosodic,Lin_2001}. Meanwhile, this mapping is far from a rigid linear template; it is dynamically modulated by multiple factors, including rhythmic constraints, information focus, and constituent length.

However, for Second Language (L2) learners, acquiring these mapping rules presents a persistent challenge. The \textit{Interface Hypothesis} posits that linguistic properties involving the interface between syntax and other cognitive systems—such as prosody—are particularly vulnerable to instability and fossilization, even at high proficiency levels \cite{interface2011,Goad2004,Chen_Mo_2013}. A critical unresolved issue in L2 prosody is the relationship between fluency and structural accuracy. While studies indicate that high-proficiency learners demonstrate superior segmentation abilities compared to beginners~\cite{Chen2022,Wang_Bing_2015_PhD}, it remains controversial whether this improvement reflects a true acquisition of native-like phrasing. As learners advance, they typically increase their speech rate and reduce prosodic boundary frequency. However, prior comparisons suggest that learner boundary hierarchies (e.g., placing breaks at VOB vs. SBV; see Table 1 for abbreviations) often deviate from the native norm even at advanced stages~\cite{Gao_Wang_2019,Yu_Danwei_2011_PhD}. Do advanced learners acquire the correct syntax-prosody mapping, or do they simply learn to speak longer by ignoring structural breaks?

Existing research often relies on surface-level metrics, such as pause duration or broad boundary frequency counts, which show that L2 learners generally produce more pauses than natives~\cite{Chen_2017,Song_PhD}. However, these metrics fail to capture the alignment between specific syntactic relations and prosodic events. \textit{Dependency Grammar} explicitly links head words to their dependents (e.g., a verb governing its object) describing the semantic dependencies between words~\cite{Liu2009,lucien1959elements,1987Dependency,2007Language}. This study addresses these gaps by conducting a comparative analysis of native Mandarin speakers and Vietnamese learners using the BLCU-SAIT speech corpus~\cite{WangWei2019} annotated with dependency guidelines~\cite{Che2021}. In this study, we compare native Mandarin speakers with Vietnamese learners stratified by proficiency. We test the hypothesis that advanced learners may achieve an "Illusion of Fluency": converging to native standards in boundary quantity, while diverging significantly in structural quality. Specifically, we investigate whether learners restructure their interlanguage prosody to match the native Subject-Predicate divide, or if they fossilize on a non-native linear grouping strategy.

\section{Methods}
\subsection{The Corpus and Participants}
This study uses data from the BLCU-SAIT corpus, a multi-modal interlanguage speech dataset designed for non-native Chinese phonetics research and teaching~\cite{WangWei2019,WangWei2020}. While the full corpus comprises approximately 21 hours of speech from 19 L1 backgrounds, the present analysis focuses exclusively on a subset of Vietnamese learners. 

The participant pool consists of 67 native Mandarin speakers (CN; 27 males, 40 females; mean age = 22.5, SD = 2.1) from mainland China, serving as the baseline control group. The experimental group comprises 67 Vietnamese learners, stratified by proficiency based on the Hanyu Shuiping Kaoshi (HSK: Chinese Proficiency Test):
\textbf{High-proficiency (VNH):} 38 participants (14 males, 24 females; mean age = 23.1, SD = 2.3). Criteria: HSK 5--6 (advanced/near-native) with high fluency ($<$10 hesitations per session).
\textbf{Low-proficiency  (VNL):} 29 participants (15 males, 14 females; mean age = 22.8, SD = 2.0). Criteria: HSK 2--4 with noticeable disfluencies.
All Vietnamese participants were recruited from Chinese language programs, balanced for gender and proficiency. The final dataset comprises 13,802 utterances (103 sentences $\times$ 134 speakers). The Vietnamese cohort was selected due to its large sample size and proficiency stratification within the BLCU-SAIT corpus, providing robust data for tracking acquisition trajectories.

\subsection{Annotation of Prosodic Boundaries and Syntax}
Prosodic boundaries used C-ToBI system(Chinese-Tones and Break Indices) ~\cite{Li2002,Li2021}, categorizing breaks B0-B4 by juncture size and acoustics (duration, pitch reset). 
Annotated by trained undergraduate linguistics majors (Cohen's kappa=0.82); graduate supervisor resolved disagreements. This study focuses on following hierarchies:B1 (Prosodic Words), B2 (Minor Prosodic Phrases), B3 (Major Prosodic Phrases).
B4 marks the Intonational Phrase (IP) boundary, which is typically sentence-final. Given the corpus's single-sentence structure, the B4 boundary was not sufficiently distinguished and is thus omitted. This study focuses on B1–B3 prosodic boundaries.

This study focuses on the following hierarchies:

\textbf{B1:} Prosodic Word (PW)

\textbf{B2:} Minor Prosodic Phrase (mPP)

\textbf{B3:} Major Prosodic Phrase (MPH)

B4 (Intonational Phrase) boundaries, typically marking sentence-final positions, were excluded as the corpus consists of single isolated sentences.

Syntactic annotation followed the dependency grammar guidelines of the CIR-CTB (Research Center for Information Retrieval Chinese Treebank)~\cite{liu2006,Che2021}.  Continuous sentences are divided into discrete words, which are linked by dependency relations. Among these, the verb is the core of the sentence, not dominated by any other elements, with all dominated elements being subordinate to their dominators through a dependency relationship. This dominance relationship reflects the semantic relations between units. Sentences were tokenized and linked via 15 dependency relations (e.g., SBV: Subject-Verb; VOB: Verb-Object; see Table~\ref{tab:hit-dependency-tags}). The process employed a semi-automatic approach: initial parsing via the HIT-LTP(Harbin Institute of Technology's Language Technology Platform) platform~\cite{Che2021} was followed by manual correction by three linguistics graduates to ensure semantic accuracy.

\begin{table}[h]
    \centering
    \vspace{-4mm}
    \setlength{\tabcolsep}{3pt}
    \footnotesize
    \caption{Harbin Institute of Technology Dependency Parsing Guidelines}
    \label{tab:hit-dependency-tags}   
    \begin{tabular}{>{\raggedright\arraybackslash}p{0.5cm} 
                     >{\raggedright\arraybackslash}p{1.7cm} 
                     >{\raggedright\arraybackslash}p{5.2cm}}
    \toprule
    \textbf{Tag} & \textbf{Description} & \textbf{Example} \\
    \midrule
     SBV & subject-verb          & I give her a bunch of flowers (I $\leftarrow$ give) \\
     VOB & verb-object           & I give her a bunch of flowers (give $\rightarrow$ flowers) \\
     IOB & indirect object       & I give her a bunch of flowers (give $\rightarrow$ her) \\
     FOB & fronting object       & He reads any book (book $\leftarrow$ read) \\
     ATT & attribute             & Red apple (red $\leftarrow$ apple) \\
     ADV & adverbial             & Very beautiful (very $\leftarrow$ beautiful) \\
     CMP & complement            & Finished the homework (do $\rightarrow$ finished) \\
     HED & head                  & Refers to the core of the entire sentence \\
    \bottomrule
    \end{tabular}
    
    \vspace{1mm}
    \footnotesize 
    Complete guidelines available at \url{http://ltp.ai/}
    \vspace{-0.6cm}
\end{table}

\section{Result}
\subsection{The Production Quantity of Prosodic Boundaries}
We first examined the density of prosodic boundaries at three hierarchical levels: Prosodic Words (B1), Minor Phrases (B2), and Major Phrases (B3). Descriptive statistics (Table \ref{tab:boundaries}) indicate substantial group differences.

Given the count nature of the data and observed overdispersion, we fitted a Negative Binomial Regression model to analyze the main and interaction effects of Speaker Group and Boundary Type. The model summary is presented in Table \ref{tab:nb_regression}.

\begin{table}[ht]
\setcounter{table}{1}
\setlength{\tabcolsep}{3pt}
\centering
{\footnotesize
\caption{Number of prosodic boundaries}
\begin{tabularx}{0.45\textwidth}{ >{\centering\arraybackslash}X| >{\centering\arraybackslash}X>{\centering\arraybackslash}X}
\toprule
\textbf{Boundary Type} & \textbf{Speaker Type} & \textbf{No.of Boundary} \\
\midrule
\multirow{3}{*}{Level:B1} & CN group & 225 \\
                    & VNH group& 190 \\
                    & VNL group& 341 \\
\midrule
\multirow{3}{*}{Level:B2} & CN group & 155 \\
                    & VNH group& 104 \\
                    & VNL group& 173 \\
\midrule
\multirow{3}{*}{Level:B3} & CN group& 70 \\
                    & VNH group & 80 \\
                    & VNL group& 124 \\
\bottomrule
\end{tabularx}
\label{tab:boundaries}
}
  \vspace{-0.5cm}
\end{table}

\begin{table}[ht]
\centering
\caption{Negative binomial regression model results. Significance levels: *** for p-value \textless0.001, ** for p-value \textless0.01, * for p-value \textless0.05, . for p-value \textless0.1, and ns for p-value $\geq$ 0.1.}
\begin{tabularx}{0.5\textwidth}{>{\centering\arraybackslash}X|>{\centering\arraybackslash}X>{\centering\arraybackslash}X>{\centering\arraybackslash}X>{\centering\arraybackslash}X}
\toprule
\textbf{Contrast} & \textbf{Estimate} & \textbf{Std. Error} & \textbf{z value} & \textbf{p-value} \\
\midrule
(Intercept) & 5.50472 & 0.06528 & 84.325 & *** \\
VNH & -0.12333 & 0.10862 & -1.136 & * \\
VNL & 0.32677 & 0.11290 & 2.894 & * \\
B2 & -0.41671 & 0.09251 & -4.505 & *** \\
B3 & -1.25623 & 0.09324 & -13.473 & *** \\
VNH:B2 & -0.32482 & 0.15420 & -2.106 & ** \\
VNL:B2 & -0.26341 & 0.16007 & -1.646 & . \\
VNH:B3 & 0.22815 & 0.15494 & 1.472 & . \\
VNL:B3 & 0.24179 & 0.16073 & 1.504 & *** \\
\bottomrule
\end{tabularx}
\label{tab:nb_regression}
  \vspace{-0.3cm}
\end{table}

\begin{figure}[ht]   
  \centering
  \includegraphics[width= \linewidth]{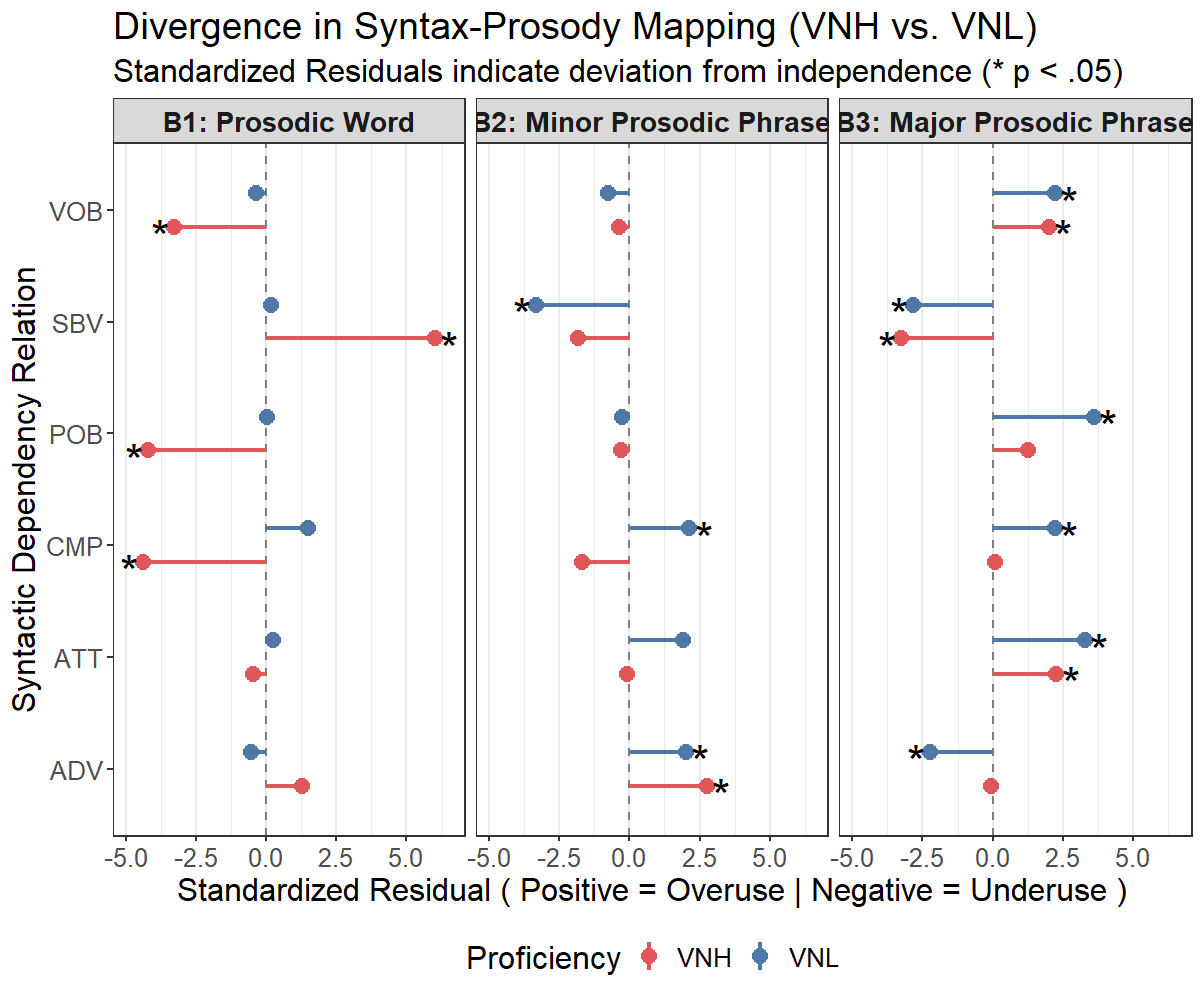}
  \caption{Deviation in Dependency Relation Frequency.( The zero line represents the CN as baseline. The bar length denotes the degree of deviation. Positive = Over-segmentation, Negative = Under-segmentation.)}
  \label{fig:diff_plot}
    \vspace{-0.4cm}
\end{figure}

The regression revealed significant interaction effects between Group and Boundary Type ($p < 0.01$). To interpret these interactions, we conducted post-hoc pairwise comparisons using estimated marginal means with Bonferroni correction. 
The analysis yielded two distinct patterns:

Global Over-Segmentation in VNL: The Low-Proficiency group ($\text{VNL}$) consistently produced a significantly higher number of boundaries than the native baseline ($\text{CN}$) across all tiers. This difference was most pronounced at the Major Phrase level ($p < 0.001$). Beginners appear to insert frequent breaks to manage high cognitive load during L2 processing~\cite{li2015}.

Tier-Specific Strategy in VNH: The High-Proficiency group ($\text{VNH}$) displayed a non-linear trajectory. At lower tiers, they produced significantly fewer boundaries than native speakers (B1: $p = 0.05$; B2: $p < 0.001$), reducing segmentation by up to 33\% at B2. However, at the highest tier ($\text{B3}$), while the regression model suggested a trend ($p<0.1$), strict post-hoc analysis confirmed no significant difference ($p=1.00$).  

This finding is critical: while VNH learners seem to have achieved "quantitative fluency" (native-like pause density) at the major phrase level, their reduction of lower-level boundaries suggests a trade-off strategy aimed at producing longer continuous runs of speech. Whether this quantitative convergence reflects accurate structural phrasing will be examined in the next section.

\subsection{Analysis of Syntax-Prosody Mapping Divergence}
To investigate whether L2 proficiency modulates the mapping mechanism between syntactic dependencies and prosodic boundaries, we performed Chi-squared ($\chi^2$) tests of independence at three prosodic levels. The results revealed a significant dependence between speaker group and syntactic distribution across all levels (all $p < 0.001$).

To quantify the specific direction and magnitude of this divergence, we calculated Standardized Residuals (StdRes) and visualized the results in Figure \ref{fig:diff_plot}, with numerical summaries provided in Table \ref{tab:residuals}. In the figure, the Zero Line represents the baseline(CN) of statistical independence, the Bar Length (Magnitude) denotes the degree of deviation, and Asterisks (*) mark statistical significance ($|StdRes| > 1.96$).

Overall, Figure \ref{fig:diff_plot} reveals a critical characteristic of parallelism: for the majority of syntactic relations, the direction of residuals for the Low-Proficiency (VNL, blue line) and High-Proficiency (VNH, red line) groups remains consistent. This directional consistency suggests that learners across proficiency levels share a similar Interlanguage Prosodic Grammar, rather than producing random errors. However, the differences in magnitude between groups reveal specific strategic adjustments at different hierarchical levels.

\begin{table}[ht]
\centering
\caption{Standardized Residuals of Syntactic Relations across Prosodic Boundaries. Values with absolute magnitude $> 1.96$ (bold) indicate significant deviation from independence.}
\label{tab:residuals}
\begin{tabular}{l c c c c}
\toprule
\textbf{Relation} & \textbf{Level} & \textbf{CN (Res)} & \textbf{VNL (Res)} & \textbf{VNH (Res)} \\
\midrule
ADV & B1 & -0.45 & -0.55 & 1.27 \\
ATT & B1 & 0.13 & 0.25 & -0.45 \\
CMP & B1 & 1.78 & 1.49 & \textbf{-4.40} \\
POB & B1 & \textbf{2.42} & 0.05 & \textbf{-4.20} \\
SBV & B1 & \textbf{-3.60} & 0.18 & \textbf{6.02} \\
VOB & B1 & \textbf{2.11} & -0.37 & \textbf{-3.30} \\
\midrule
ADV & B2 & \textbf{-2.29} & \textbf{2.02} & \textbf{2.75} \\
ATT & B2 & -0.87 & 1.92 & -0.08 \\
CMP & B2 & -0.19 & \textbf{2.14} & -1.68 \\
POB & B2 & 0.26 & -0.26 & -0.28 \\
SBV & B2 & \textbf{2.47} & \textbf{-3.33} & -1.83 \\
VOB & B2 & 0.55 & -0.77 & -0.38 \\
\midrule
ADV & B3 & 1.95 & \textbf{-2.23} & -0.06 \\
ATT & B3 & \textbf{-4.38} & \textbf{3.29} & \textbf{2.28} \\
CMP & B3 & -1.98 & \textbf{2.23} & 0.09 \\
POB & B3 & \textbf{-3.95} & \textbf{3.61} & 1.25 \\
SBV & B3 & \textbf{4.68} & \textbf{-2.85} & \textbf{-3.28} \\
VOB & B3 & \textbf{-3.27} & \textbf{2.23} & \textbf{1.99} \\
\bottomrule
\end{tabular}
\end{table}
\subsubsection{B1 Level: Prosodic Demotion Strategy}
At the Prosodic Word level, native speakers exhibited a canonical (Subject)|(Verb-Object) pattern, significantly inhibiting boundaries at the Subject-Verb (SBV) interface ($StdRes = -3.60$) while facilitating them at the Verb-Object (VOB) interface ($StdRes = +2.11$), reflecting lexical-level cohesion within the predicate.

However, the VNH group displayed an extreme anomalous pattern at this level. \textbf{Significance}: VNH speakers showed a highly significant positive residual at SBV ($StdRes = +6.02$, represented by the longest positive red bar in the B1 panel of Figure \ref{fig:diff_plot}), whereas the VNL group showed a non-significant deviation ($+0.18$). \textbf{Interpretation}: This reveals a strategy of "Prosodic Demotion": advanced learners perceive the syntactic boundary after the subject, but relegate the structural break to a weak, word-level juncture (B1) rather than the appropriate higher-level phrase break. While this strategy may improve perceived fluency by by skipping pauses, it compromises the clarity of the syntactic hierarchy.

\subsubsection{B2 Level: The Transitional ZoneThe Minor Phrase level functions as a transition from lexical to phrasal prosody. }
Native speakers began to show a facilitation preference at the SBV position ($StdRes = +2.47$). \textbf{Magnitude Comparison:} Both learner groups exhibited under-segmentation (inhibition) at this position. While the VNL group showed severe inhibition ($StdRes = -3.33$), the VNH group showed a reduced magnitude of deviation ($StdRes = -1.83$), indicating a trend towards the native baseline. However, this did not translate into accurate segmentation, but rather foreshadowed the displacement of non-canonical patterns to other levels.

\subsubsection{B3 Level: Fossilization of Non-Native Phrasing}
At the critical Major Phrase level, the data provides the strongest evidence for the Interface Fossilization Hypothesis. Native speakers established a clear (Subject)-(Verb- object) structure (SBV: $+4.68$; VOB: $-3.27$). In contrast, learner groups exhibited an inverse (Subject-Verb)-(Object) mapping pattern.Figure \ref{fig:diff_plot} (B3 Panel) clearly demonstrates that this deviation does not attenuate with proficiency, but rather entrenches at key nodes:

\textbf{SBV Demotion}: The Low-Proficiency group (VNL) already reduced the native SBV boundary ($-2.85$). However, the High-Proficiency group (VNH) showed an even stronger deviation ($-3.28$). This indicates that as proficiency improves, learners do not correct this pattern; instead, the strategy of skipping the post-subject pause becomes more systematic and entrenched..

\textbf{VOB Promotion}: Both learner groups significantly inserted unnatural major phrase (B3) between the verb and object (VNL: $+2.23$; VNH: $+1.99$), with both deviations being statistically significant.
At the verb-object spot, native speakers mainly use B1 to keep words together (StdRes = +2.11). In contrast, the VNH group shows a big shortage at this B1 level (StdRes = -3.30) and too much at the major phrase level (B3, StdRes = +1.99).
This means the VNH learn have trouble making the small breaks needed to link verb and object. Instead, when they do break, they wrongly pick a strong phrase break, which splits the meaning of the verb part.

The non-convergence in magnitude at B3—where the VNH group deviates largely from the native norm—suggests that the increased fluency of VNH learners stems not from accurate syntactic parsing, but from the fossilization of an incorrect (Subject-verb)-(Object) prosodic template. This finding challenges the assumption that proficiency guarantees accuracy, confirming the resistance of the syntax-prosody interface to restructuring in L2 acquisition.

\section{Discussion}
This study integrates quantitative boundary analysis with dependency mapping analysis to investigate the acquisition trajectory of the syntax-prosody interface in L2 Mandarin. The combined results reveal a complex non-linear trajectory: while high-proficiency learners (VNH) achieve native-like boundary quantity at the B3, their structural mapping remains fundamentally non-native and fossilized.

\subsection{The Illusion of Fluency: Quantity Convergence vs. Quality Divergence}
Experiment 1 demonstrated that VNH learners have successfully converged to the native baseline in terms of boundary density at the Major Phrase (B3) level ($p = 1.00$). This contrasts sharply with the significant over-segmentation observed in low-proficiency (VNL) learners ($p < 0.001$). This reduction in pause frequency suggests that advanced learners have achieved a high level of "quantitative fluency"—the ability to produce longer continuous runs of speech—thereby overcoming the fragmentation characterizing lower-proficiency speech.

However, the standardized residual analysis in Experiment 2 reveals that this quantitative convergence masks a qualitative divergence. While VNH speakers produce the correct number of pauses, they place them in the wrong syntactic locations. Specifically, VNH speakers significantly under-segment at the native-canonical Subject-Verb (SBV) interface ($StdRes = -3.28$) while over-segmenting at the non-native Verb-Object (VOB) interface ($StdRes = +1.99$). This indicates that VNH learners have acquired the surface rhythm of Mandarin (when to stop) without acquiring the underlying syntax-prosody grammar (where to stop). This "pseudo-fluency" highlights the limitations of evaluating L2 prosody based solely on boundary count metrics.

\subsection{ The Prosodic Demotion Strategy}
Synthesizing findings from both experiments, we propose the "Prosodic Demotion" hypothesis to explain how advanced learners maintain long phrase while sacrificing structural clarity. The evidence for this strategy is twofold:At the Major Phrase level (B3), VNH learners significantly reduces the SBV boundary ($StdRes = -3.28$), failing to mark the subject with a major phrase boundary. Conversely, at the Prosodic Word level (B1), VNH learners exhibit an extreme spike in segmentation at the same SBV position ($StdRes = +6.02$).This stark contrast suggests that advanced learners do not suffer from "boundary blindness"; rather, they employ a strategy of demotion. To accommodate the pressure for continuous speech flow, they relegate the structural break—which requires a B3 boundary—to a weak, B1, such as mere syllable lengthening or a micro-pause. This strategy satisfies the constraint for boundary reduction, it makes the speech continuous.

\subsection{Fossilization and the Linear Adjacency Effect}
Our findings provide robust support for the Interface Hypothesis \cite{interface2011}, confirming that the syntax-prosody interface is resistant to restructuring even at advanced levels. Crucially, the data exhibits an "Inverse Magnitude" at the B3 level: the negative deviation at the SBV interface is stronger in the advanced VNH group ($StdRes = -3.28$) than in the lower-proficiency VNL group ($StdRes = -2.85$). The pattern that points toward fossilization rather than gradual acquisition.

If the non-native phrasing were merely a performance error, we would expect the deviation to decrease as proficiency increases. Contrary to this prediction, the inhibition of the canonical SBV boundary is stronger in the advanced VNH group than in the lower-proficiency VNL group. And this pattern cannot be attributed to mere performance errors driven by maintain long phrasal output. If articulation speed were the sole driver, we would expect a uniform reduction in boundary strength across all syntactic heads. Instead, the specific reduction of SBV combined with the facilitation of VOB confirms that this is a fossilized structure into a (Subject-Verb)-(object) template. This suggests that as learners gain fluency, the non-native strategy of grouping the Subject and Verb via Linear Adjacency becomes more, not less, entrenched. The "Subject-Verb" cohesion—likely a processing strategy to secure the semantic core of the sentence rapidly—stabilizes into a fossilized prosodic template, a pattern potentially reinforced by L1 Vietnamese prosodic structures which warrants further cross-linguistic investigation.

\section{Conclusion}
This study challenges the assumption that syntactic proficiency guarantees prosodic competence by triangulating boundary quantity with syntactic distribution. We draw three major conclusions:
\textbf{Non-Linear Acquisition:} L2 prosodic development does not follow a linear trajectory toward the native norm. While boundary quantity may converge to native levels, the underlying mapping rules (quality) can diverge. We observe a complex trade-off between metric fluency and structural accuracy.
\textbf{Mechanism of Demotion:} Advanced learners resolve the conflict between continuity and structural marking via "Prosodic Demotion." By demoting essential structural boundaries (e.g., Post-Subject breaks) to weak lexical junctures, they prioritize linear flow over hierarchical clarity.
\textbf{Evidence of Fossilization:} The non-native (Subject-Verb)-(Object) phrasing pattern becomes fossilized at advanced levels. The finding that high-proficiency learners deviate further from the native structural baseline than beginners highlights the resistance of the syntax-prosody interface to standard acquisition.

This study demonstrates that L2 prosodic acquisition is non-linear and distinct from syntactic development. While Vietnamese learners successfully acquire Mandarin SVO syntax, they map it onto a fossilized prosodic template that groups Subject and Verb. This divergence is most pronounced in advanced learners, highlighting the resistance of the syntax-prosody interface to restructuring. 
\bibliographystyle{IEEEtran}

\bibliography{mybib}

\end{document}